%% file: acl_latex.tex
\pdfoutput=1

\documentclass[11pt]{article}

\usepackage[final]{acl}

\usepackage{times}
\usepackage{latexsym}

\usepackage[T1]{fontenc}

\usepackage[utf8]{inputenc}

\usepackage{microtype}

\usepackage{inconsolata}

\usepackage{graphicx}

\usepackage{hyperref}
\usepackage{multirow}
\usepackage{tabularx}
\usepackage{booktabs}
\usepackage{amsmath}
\usepackage{subcaption}

\newcommand{\anonymize}[1]{#1}

\newcommand{\filluptopage}[1]{%
  \clearpage
  \loop\ifnum\value{page}<#1\relax
    \null\clearpage
  \repeat
}

\title{Classifying long legal documents using short random chunks}

\author{Luis Adrián Cabrera-Diego \\
  Jus Mundi / 30 Rue de Lisbonne, 75008 Paris, France \\
  \texttt{a.cabrera@jusmundi.com}}

\begin{document}
\maketitle
\begin{abstract}
    Classifying legal documents is a challenge, besides their specialized vocabulary, sometimes they can be very long. This means that feeding full documents to a Transformers-based models for classification might be impossible, expensive or slow. Thus, we present a legal document classifier  based on DeBERTa V3 and a LSTM, that uses as input a collection of 48 randomly-selected short chunks (max 128 tokens). Besides, we present its deployment pipeline using Temporal, a durable execution solution, which allow us to have a reliable and robust processing workflow. The best model had a weighted F-score of 0.898, while the pipeline running on CPU had a processing median time of 498 seconds per 100 files.
\end{abstract}

\section{Introduction}

    \textit{Legal AI} is the use of artificial intelligence technologies, to help legal professionals in their heavy and redundant tasks \cite{zhong-etal-2020-nlp}. And, while Legal AI is not new \cite{dale_law_2019}, processing legal documents is challenging.
    
    The main challenge is that legal documents are diverse, not only in their length, but also in their vocabulary, structure, subjectivity and scope \cite{mitchell_document_2014, trautmann_large_2023}. And while the two latter characteristics can be (partially) minimized by training tools using specialized corpora \cite{chalkidis_legal-bert_2020}, the first characteristic, i.e. length, cannot be. For instance, the longer the document, the harder to keep the correct contexts that are actually relevant for a task \cite{wagh_comparative_2021}. As well, when using Transformed-based technologies, such as \textit{BERT} \cite{devlin_bert_2019}, the memory consumption will explode the longer the input \cite{vaswani_attention_2017}. And while there are now \textit{Large Language Models (LLM)}, such as \textit{GPT}\footnote{https://openai.com/}, that can process thousands of tokens, their use can be expensive, or can have risks \cite{karla_grossenbacher_employers_2023,cassandre_coyer_legal_2024,fields_survey_2024}.

    Therefore, we present a classifier capable of processing long legal documents, and can be deployed within in-house CPU servers. To achieve this, we have created a document classifier using \textit{DeBERTA V3} \cite{he_debertav3_2021} and a \textit{LSTM}, that uses a collection of 48 randomly-selected short chunks (the maximum size of the chunks is 128 tokens). As well, we describe how we used \textit{Temporal}\footnote{\href{https://temporal.io/}{https://temporal.io/}} to create durable workflows and to focus on the delivery of our classifier and not how to deal with the workflows' execution.

    The proposed model, has been trained and tested using a large multilingual collection of legal documents of different lengths and that covers 18 classes. The classifier has a median weighted F-score of 0.898 and the median time for processing files using Temporal is 498 seconds per 100 files.

\section{Related Work}
\label{sec: RelatedWork}

    Most of the works related to the classification of long documents rely on the splitting of documents. For instance,  \citet{pappagari_hierarchical_2019} segment a long document into smaller chunks of 200 tokens, feed them into a \textit{BERT} \cite{devlin_bert_2019} model, and propagate them into either an LSTM or a transformer layer. \textit{CogLXT} \cite{ding_cogltx_2020} is a classifier that uses only key sentences as input; these are obtained using a model trained as a judge. \citet{park_efficient_2022} presented two baselines using BERT where relevant sentences, determined by \textit{TextRank} \cite{mihalcea-tarau-2004-textrank}, or randomly selected ones, are concatenated to the first 512 tokens of a document.
    
    

    Others works have explored how to increase the input size of Transformer-based models. The best example is \textit{Longformer} \cite{beltagy_longformer_2020}, a model that is capable of processing up-to 4,096 tokens by using a windowed local-context self-attention and a global attention. Nonetheless, this kind of models tend to suffer from large memory consumption and long processing time \cite{park_efficient_2022}.

    On the legal domain, we can highlight the following works. \citet{chalkidis_large-scale_2019} classified documents by training legal \textit{Doc2Vec} embeddings \cite{pmlr-v32-le14} and feeding them into a BiGRU with a \textit{Label-Wise Attention Network} \cite{mullenbach-etal-2018-explainable}. Similarly, \citet{wan_long-length_2019} trained legal Doc2Vec embeddings, but they fed them into a BiLSTM, with chunk attention layer.
    

    \textit{LegalDB} \cite{bambroo_legaldb_2021} and \textit{Lawformer} \cite{xiao_lawformer_2021} converted respectively \textit{DistillBERT} \cite{sanh2020distilbertdistilledversionbert} and a \textit{Chinese RoBERTa model} \cite{9599397} into a Longformer model. Both models were pre-trained using legal corpora. Similarly,  \citet{mamakas_processing_2022} converted \textit{LegalBERT} \cite{chalkidis_legal-bert_2020} into a legal hierarchical BERT, and a legal Longformer model which can process up to 8,192 tokens.
    
    

    \textit{D2GCFL} \cite{wang_d2gclf_2022}, is a legal document classifier that extracts relations and represent them into four graphs, which are fed into a graph attention network. \citet{li_dont_2023} split legal contracts into smaller chunks, which are filtered using a logistic regression model and then, these are fed into a fine-tuned \textit{RoBERTa} \cite{liu_roberta_2019}.
    

    \citet{trautmann_large_2023} uses prompt chaining for classifying legal documents. The author summarize a legal document, by creating recursively summaries of smaller chunks. The summary is sent to an LLM with a few-shot prompt. The prompt has examples that were generated using the summary approach.

\section{Work's Scope}
\label{sec: Scope}

    This work presents a tool developed by and for \anonymize{\textit{Jus Mundi}}\footnote{\anonymize{\href{https://jusmundi.com/}{https://jusmundi.com/}}}, a legal-tech company, with the purpose of solving an internal need. In other words, \anonymize{Jus Mundi} wanted to create a legal document classifier, that can be used in multiple services and tools, internal and external, with the following characteristics:
    \begin{itemize}
        \item Privacy-focused: Many LLMs use the data feed into the services to train future models, which, in the legal domain, might pose a privacy risk.
        \item Quick: The classifier needs to be used in some real-time tools, thus, speed is an essential aspect to take into account.
        \item Simple: Simple models are easier to maintain and to retrain if necessary. In other words, complex architectures, where multiples models are necessary, such as \textit{CogLXT} \cite{ding_cogltx_2020}, are difficult to maintain due to their increased technical debt. As well, classifiers based on prompts are sometimes hard to maintain if new classes are added or the LLM model is updated.
        \item Not expensive: Some models found in the literature need expensive hardware, i.e. big GPUs with large amounts of VRAM, that for production environments might be too expensive to run on the long term. Thus, it is essential to create a solution that could run the inference task in servers with only CPUs or small GPUs.
    \end{itemize}

    For achieving this objective, we decided to explore the following hypothesis: \textit{It is possible to classify (long) legal documents using small and randomly-selected portions of text}. This hypothesis is based on observations and results presented in previous research works. For instance, \citet{ding_cogltx_2020,park_efficient_2022,li_dont_2023} noticed that it is not necessary to pass the full document to classify it correctly. Moreover, a random selection of passages perform in some cases equally or even better than complex models such as those based on Longformer, CogLTX, or ToBERT \cite{park_efficient_2022}.

\section{Methodology}
\label{sec: Methodology}

\begin{figure}[t!]
    	\centering
    	\includegraphics[width=0.37\columnwidth]{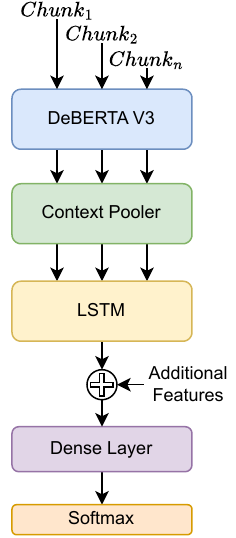}
    	\caption{Proposed architecture of the classifier.}
    	\label{fig:architecture}
    \end{figure}

    We propose an architecture based on the \textit{multilingual DeBERTa V3} \cite{he_debertav3_2021} model\footnote{\href{https://huggingface.co/microsoft/mdeberta-v3-base}{https://huggingface.co/microsoft/mdeberta-v3-base}} and a LSTM. In detail, first, a collection of chunks, from a document, are passed through DeBERTa V3 to get their contextual embeddings. Then, the first token of each embedding, i.e. the \textit{[CLS]} token, is passed through a dense layer and a GELU activation, to get a context pool.\footnote{This is called a Context Pooler Layer, see:\\ \href{https://deberta.readthedocs.io/en/latest/_modules/DeBERTa/deberta/pooling.html\#ContextPooler}{https://deberta.readthedocs.io/}} After that, the collection of context pools are fed into the LSTM. If available, additional features will be concatenated to the final state of the LSTM. The final state will be fed then to a dense layer and a Softmax activation for determining the class of the document. In Figure \ref{fig:architecture}, we present the architecture.

    As indicated in Section~\ref{sec: Scope}, rather than feeding a full document, either as whole or split into small chunks, to the neural network, we decided to randomly sample chunks. Therefore, we explored three sample sizes, 20, 48 and 62. The sizes 20 and 62 were selected according to the paragraphs' quartiles of the corpus (see Table~\ref{table:corpus}); the size 48 was randomly chosen.\footnote{Although we tried to explore, larger contexts, up to samples of 160 chunks, their training was complex. Those that we partially managed to train (80 and 120), were slower, used more memory, and were not better than those presented here.}
    
    To create the chunks, we split the documents into paragraphs\footnote{Our data was originally in HTML, thus we used tags such as <p> (paragraph) and <li> (listing) for splitting the texts.}. Then, each paragraph is tokenized and encoded with DeBERTa V3 tokenizer. If the generated chunk surpasses 128 tokens, they are sub-split with an overlapping stride of 16 tokens. The sampling is done over all the document's encoded chunks, and they are fed into the neural network following the document's order.

    Besides, we explored 3 types of document length representations as additional features\footnote{According to our legal experts, the document length could be a good discriminator of the classes.}: number of characters ($n_c$), number of paragraphs ($n_p$), 
    and approximate number of pages ($a_{pp} = n_c / 1,800$)\footnote{We do not have the original number of pages for all documents, this is why it is approximated.}. To keep these features in a close range of values, all of them were represented using the natural logarithm.

    In Table~\ref{tab:hyperparams}, we present the hyperparameters used for training the classifier. Moreover, it should be noted that during the training process, the sampling process is done at the beginning of each epoch. In other words, in each epoch, we feed the neural network a collection of documents that has different chunks. The goal was to make more robust the training of the document classifier by providing different portions of text that could occur in a legal document.

\input{latex/Tables/hyperparams}

\section{Data}
\label{sec: Data}

    We use a proprietary multilingual legal corpus related to the domain of legal arbitration and covering 25 languages\footnote{Certain documents have versions in multiple languages. For those cases, we only considered one language, giving priority to those different from English, French, Spanish and Portuguese, since these were the most frequent ones: 9,378, 1,602, 998, and 255 documents respectively.}.
    
    All the documents were manually classified by legal experts into 18 classes throughout multiple years. In this work, we divided this corpus into 3 sets, train (80\%), dev (10\%), and test (10\%); all the data is represented using JSON line files (JSONL). 
    In Table~\ref{table:corpus}, we present the corpus' statistics.

    It should be seen that the approximate median number of pages of all the corpus is 7.3. As well, that the longest documents in the corpus are those belonging to the class \textit{Expert opinion} (the median is 38.8 pages), while the shortest ones are those belonging to the class \textit{Other} (the median is 1 page). These figures were calculated by diving the median number of characters by 1,800 (number of characters in a standard page).

    \input{latex/Tables/corpus}

\section{Deployment}

    Since the goal of the classifier is to be part of a larger project that will be used along other tools, such as an OCR, and a metadata extractor, we decided to deploy it within a processing pipeline. Specifically, we used \textit{Temporal}, an open-source orchestrator that allow us creating robust and scalable workflows, in a simple way. Where we focus only on the business logic, rather than coding as well, aspects such as distribution of tasks, queue systems and state capture. Thus, in this paper, we present the current workflow created in Temporal and its deployment. However, it should be indicated that the described pipeline will change in the near future, since we will add other components that currently are being implemented.

    Temporal uses 4 basic elements: \textit{Activities} (a task or collection of simple tasks), \textit{Workflows} (a collection of Activities), \textit{Queues} (a waiting list of inputs for Activities and Workflows), and \textit{Workers} (a deployment instance that runs specific Activities and/or Workflows and that listens a specific Queue). Moreover, each Activity and Workflow process only one input at the time, but multiple Activities and Workflows can be run at the same time. This can be done either by configuring the Workers limits or by deploying more Workers.

    All the input and outputs from/for a user (either a human or another system) are communicated using Temporal's client. This client can call a specific workflow to run, with its respective input, and can either return an ID (to ask later for the status), or wait for the workflow to finish and return its output.

    \subsection{Pipeline}
    \label{sec: Pipeline}

        Our pipeline, at the current state, is composed of three Activities. The first one, $a_1$, for a given directory, it finds all the JSON files and produce a list of path files. The second one, $a_2$, it reads a JSON file and validates it. The third one, $a_3$, process the JSON file to create the input of the neural network and calls the neural network to infer the class.\footnote{We did not use a different Activity for the creation of the neural network input, since it would have been too expensive to do. In other words, we would need to convert a document (JSON) into Numpy arrays (by tokenizing and encoding it) and then into a new JSON. Then this new JSON would have to be parsed to convert it into a Torch tensor.} Furthermore, this last Activity listens to its unique Queue, $Q_c$, which is dedicated for documents ready to be classified, rather than the Queue $Q_{io}$, which is for input-output processing.
        
        These Activities are split into two Workflows. The first Workflow, $w_1$, calls $a_2$ and $a_3$; moreover, it creates batches of 10  documents to process. When a batch has been finished, if there are more documents to process, it will create a new Workflow instance\footnote{This is to avoid an explosion of the history event. See \href{Continue as new}{https://docs.temporal.io/develop/python/continue-as-new}.}. The second Workflow $w_2$ calls Activity $a_1$ and Workflow $w_1$.
        
\begin{figure}[t!]
        	\centering
        	\includegraphics[width=0.9\columnwidth]{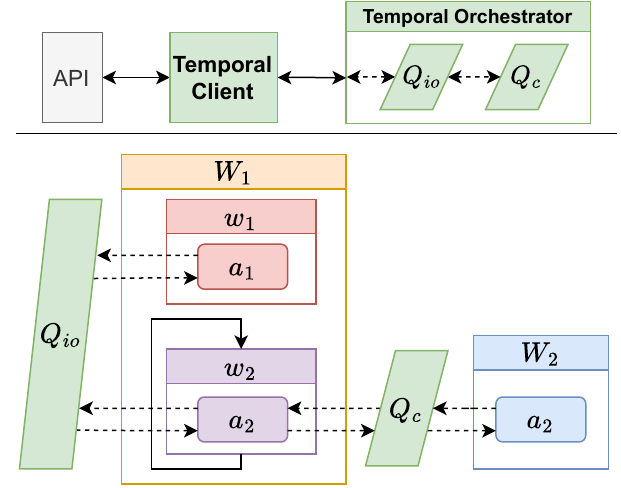}
        	\caption{Temporal's pipeline architecture. Dashed lines are indirect communication managed by Temporal.}
        	\label{fig:pipeline_architecture}
        \end{figure}

        Finally, we have designed two Workers. The Worker $W_1$ is charged of sending data to the Queue $Q_{io}$, and it manages, $a_1$ and $a_2$, plus $w_2$ and part of $w_1$. The Worker $W_2$, connects only to the Queue $Q_c$, which means that it only runs $a_3$. We can have as many instances of each worker, and all of them will be orchestrated by Temporal automatically. An API, deployed independently, sends the input to process to Temporal. In Figure~\ref{fig:pipeline_architecture}, we have a diagram of Temporal's pipeline architecture.

\section{Results and Discussion}
\label{sec: Results}

    \input{latex/Tables/results}

    We present, in Table~\ref{table:results}, the results of the top five models according to their weighted F-score over the dev set calculated during their training.\footnote{We use the weighted F-score since the corpus is imbalance. Some classes in the dev and test set have as few as 14 documents.} As well, we present the result of the smallest sampling explored. It should be noted that the results, regarding the test set (Table~\ref{table:results}), were calculated over 30 runs. The reason is that since we do at runtime (either training or inference) a sampling of the documents' paragraphs, each run will generate a different weighted F-score. Thus, in this way, we can have a better idea of the actual performance distribution for each model.

    In Table~\ref{table:results}, we can observe that, in general, the models perform within a respectable range of values over the test set, and no great variations are found. Moreover, there is no big difference between the obtained weighted F-score on the dev set and the median ($Q_2$) obtained on the test set. This means that our approach of changing each epoch the chunks sent to the neural network made ours models robust. However, the best weighted F-score in the dev set did not reflect the best weighted F-score in the test set, although this can be seen as problematic, this was expected, since during the training we only run once per epoch the evaluation on the dev set.\footnote{In the future, we will add the possibility of running multiple evaluations during the training, to be sure to select the actual best model.} Interestingly, we can also notice in Table~\ref{table:results}, that even as few as 20 chunks can be useful for classifying documents, although with a lesser performance, but still good enough for many tasks.

    We present, in Table~\ref{table:corpus_results}, the median F-scores obtained per each class in the test set over 30 runs\footnote{The same that were used for Table~\ref{table:results}.} regarding the best models found according to the dev and test sets. As we can see in Table~\ref{table:corpus_results}, most of the median F-scores are greater than 0.800, few are the exceptions, such as \textit{Pleadings}, \textit{Notice of intent}, and \textit{Settlement agreement}. These last two classes of documents were some of the less frequent documents in the training corpus. 

    We want to highlight, that we decided to include the additional features because during the first experiments we were having issues differencing two classes, \textit{Notice of intent} and \textit{Notice of Arbitration}. Both are similar in structure, but according to our legal experts, the former tend to be shorter than the latter. Although, this is not supported according to the corpus' statistics (Table~\ref{table:corpus}), we considered that nevertheless, the additional features could improve in general the classification. Nonetheless, the addition or not of additional features did not improve the classification of \textit{Notice of intent}, which was the class that was the most incorrectly misclassified of all the dataset. While we consider that, maybe it is because \textit{Notice of intent} is one of the less frequent documents in the corpus, we ask ourselves as well whether the issue comes as well from annotation errors, which might also explain the length disagreement between the facts and the legal knowledge of our team. 

    Furthermore, it is interesting to notice, that there is not a clear relationship between low score, few documents and/or document length from comparing Table~\ref{table:corpus} and Table~\ref{table:corpus_results}. For instance, \textit{Pleadings} is a frequent class, but it had low performance. The reason is that some \textit{Notice of request} and \textit{Other requests} are sometimes considered as \textit{Pleadings}. As well, in occasions \textit{Expert opinions} were misclassified as \textit{Witness statements}, regardless of whether we used the additional features and the fact that the former tend to be the longest documents in the corpus. Something similar happens between \textit{Settlement agreement} and \textit{Contracts}, where the latter is sometimes the class predicted for the former documents, despite their difference in length according to Table~\ref{table:corpus}.

    \input{latex/Tables/detailed_results}

\section{Deployment analysis}
\label{sec: lessons}

\begin{figure*}[t!]
        \centering
         \begin{subfigure}[b]{\textwidth}
            \centering
            \includegraphics[width=.95\textwidth]{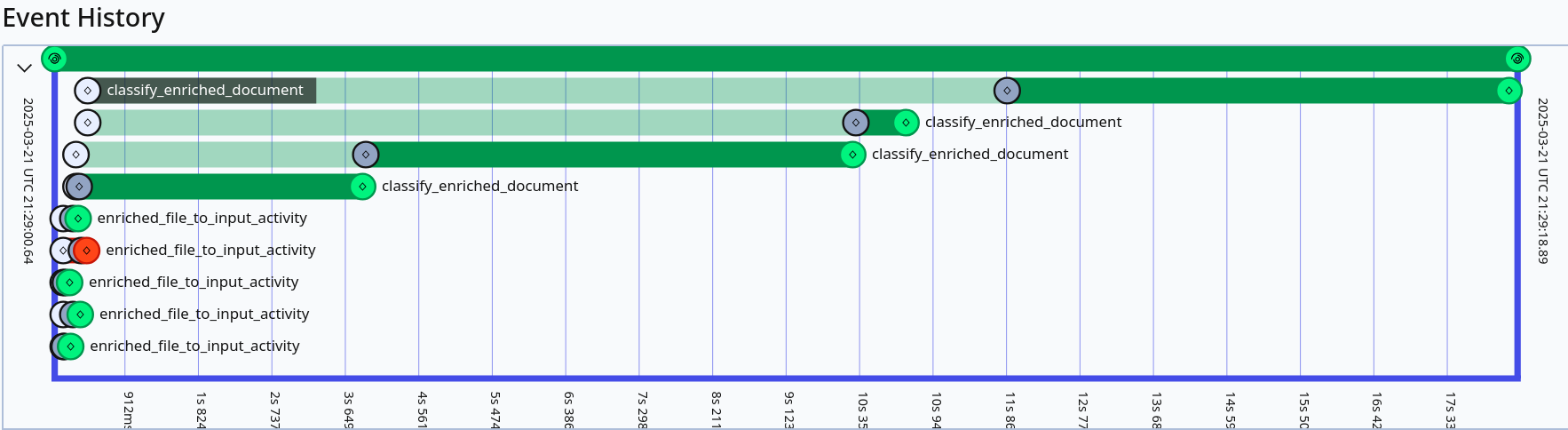}
            \caption{Event history as a diagram. The first circle indicates when the task was received in the queue; the middle one is when it started to run; the last one is when it ended. The fourth Activity, \textit{enriched\_file\_to\_input\_activity}, is red because it failed.}
        \end{subfigure}
        \begin{subfigure}[b]{\textwidth}
            \centering
            \includegraphics[width=.95\textwidth]{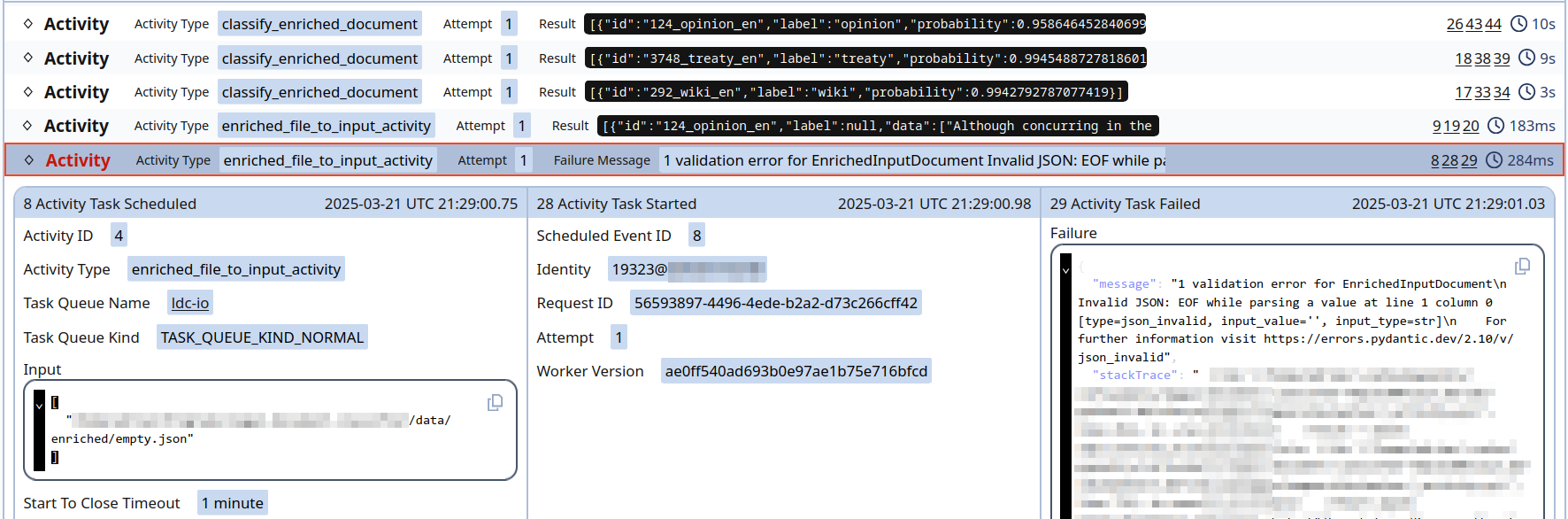}
            \caption{Event history as a collection of payloads. We can observe the failed activity and the stack trace.}
        \end{subfigure}
        \caption{Temporal's event history. Despite errors in the workflow, the pipeline continued until completed.}
        \label{fig:event_history}
    \end{figure*}

    With respect to the deployment using Temporal, we consider it a useful and interesting tools for creating pipelines. For instance, it took us around 2 hours to develop a functional pipeline locally\footnote{This does not include reading the documentation or deploying the Temporal's Orchestrator in our servers.}, and less than a week having a much more robust pipeline that we could deploy in our servers. As well, we really focused more on what we want the pipeline to do, rather than focusing on how to orchestrate asynchronous tasks, workers, queues, etc. For instance, it was very simple to indicate that if the reading or validation of a file was unsuccessful (e.g., does not exist or has the wrong format), the Activity should not be retried again, without failing the whole process (see Figure~\ref{fig:event_history}). But, if a Worker was down or stopped giving signals of being alive, the failed Activities should be retried $n$ times. Also, it is very simple to scale horizontally, we just need to deploy another Worker, and Temporal will manage the rest, e.g. sending to the Worker Activities that were waiting in the queue.

    In Temporal's interface, we can see the workflows' event history (Figure~\ref{fig:event_history}). For example, we can see how long a Workflow/Activity took to start/finish, by which worker it was done, which were the errors, and how many times were retried.\footnote{This does not replace tools such as \href{https://www.datadoghq.com/}{DataDog}, but it can help developers to better understand the pipeline.}

    Furthermore, encrypting and decrypting Temporal's payloads (messages send between Activities, Workflows and Workers), was extremely easy. The only issue we had not expected about Temporal's payloads, was their limited size. In other words, Temporal communication between workflows and activities are done using \textit{gRPC}\footnote{\href{https://grpc.io/}{https://grpc.io/}} payloads, and these are limited in size: 2MB for requests and 4MB for event history.\footnote{It is possible to increase their size, but the best practices indicate that it should be \href{https://kreya.app/blog/grpc-best-practices/\#large-messages}{avoided in languages with garbage collection}; and that the ideal size is in fact less than \href{https://github.com/grpc/grpc-java/issues/1676}{1MB}, and more around \href{https://github.com/grpc/grpc.github.io/issues/371}{16KiB to 64KiB}.} Thus, we had to change how and how many documents we processed. In other words, we decided to split the list of documents to process in batches of size 10; this splitting is done within Temporal's logic, thus, for the user there is no further task to be done. 

    We present, in Table~\ref{table:inference_time}, the time needed for running 30 pipelines using Temporal for predicting 100 documents. The 100 documents were randomly sampled each run from the test set; the goal was to explore the functioning of the pipeline over different types of documents, and in consequence different combinations of lengths. As well, to explore whether a batch of 10 documents was a good fit for Temporal's payloads.

    \input{latex/Tables/inference_time}

    It should be noted that the results presented in Table~\ref{table:inference_time} were done using a server that only has CPUs, and only one worker of each type ($W_1$ and $W_2$) was deployed. As well, the inference time comprises the reading and validation of the documents; in other words, it evaluated the full pipeline.

    We can observe, in Table~\ref{table:inference_time}, that the median time to process 100 documents is 498.080 seconds ($\sim$4.98 seconds per file). While it might not be bad for a CPU processing (see \citealp{park_efficient_2022} for other comparisons), it is certainly far from the median (calculated over more than 300 runs) of 68.002 seconds that an \textit{NVIDIA A100} took for predicting the full test set (1,229 documents).

\section{Conclusions and Future Work}

    Automatizing the classification of legal documents, while necessary for simplifying the tasks of lawyers, can be challenging. This is due to the characteristics of legal documents, such as specialized vocabulary and length.

    In this work, we introduced how \anonymize{Jus Mundi} solved an internal need by exploring the following hypothesis \textit{it is possible to classify (long) legal documents using small and randomly-selected portions of text}. This hypothesis was defined due to internal constraints, but also from observations and results that other researchers in the state-of-the-art have produced. Therefore, we presented the architecture and methodology for creating a classifier for legal documents that uses up to 48 random chunks of maximum 128 tokens. Furthermore, in this work, we explained how we deployed the classifier in our servers using Temporal, an open-source durable execution solution, for creating pipelines that could be used by our tools and services.

    The results obtained, showed that it is possible to classify documents correctly using random short chunks. Specifically, our models  can reach a median weighted F-score of 0.898 and a median speed of $\sim$4.98 seconds per file on a CPU server. As well, we noticed that the use of Temporal for creating pipelines can simplify their design, coding and deployment.

    In the future, we want to convert our current PyTorch model into a \textit{ONNX}\footnote{\href{https://onnx.ai/}{https://onnx.ai/}} one, using the \textit{ONNX Runtime}\footnote{\href{https://onnxruntime.ai}{https://onnxruntime.ai}}. The goal is to improve the inference by optimizing the model, even exploring its quantization. As well, we want to compress payloads using \textit{Zstandard}\footnote{\href{https://facebook.github.io/zstd/}{https://facebook.github.io/zstd/}}, for increasing Temporal's batch size. And at the same time, we need to improve the batching logic used in the neural network when deployed using Temporal, since right now it only processes one document at the time.

    Finally, we want to analyze deeper the misclassified documents with our legal team, in order to better understand what could be the reasons for some of the errors presented in this work. And to analyze whether the language of a document plays a role regarding the classification problems of certain documents and how it affects the global performance of our legal document classifier.


\section*{Ethical considerations}

    All the documents used for the development of the classifiers were obtained fairly. In other words, these were either written by our legal team or obtained from public sources, collaborations and through partnerships that indicated that we could use their documents for training machine learning models. As well, documents with sensitive information were previously anonymized, by experts, and we, the developers of the classifier, did not have access to the original documents. 

\section*{Acknowledgments}
    This work was possible due to the granted access of IDRIS (Institut du Développement et des Ressources en Informatique Scientifique) High-performance computing resources under the allocation 2024-AD011012667R3 made by GENCI (Grand Équipement National de Calcul Intensif).

\bibliography{custom}

\end{document}

%% file: latex/Tables/hyperparams.tex
\begin{table}[t]
	\centering
	\caption{Hyperparameters used for training the models using an NVIDIA - A100 80Gb.}
	\label{tab:hyperparams}
    \vspace{-0.3em}
    \scalebox{0.86}{
	\begin{tabular}{ll}
		\toprule
		\textbf{Hyperparameter} & \textbf{Value} \\
		\midrule
		Maximum Epochs 					& 35 \\
		Early Stop Patience 			& 5 \\
		Learning Rate 					& 2e-5 \\
		Scheduler						& Linear with warm-up \\
		Warm-up Ratio					& 0.1 \\
		Optimizer						& Lookahead \\
                                            & \cite{zhang_lookahead_2019} AdamW \\
                                            & with bias correction \\
		AdamW $\epsilon$				& $1 \times 10^{-8}$ \\
		Random Seed						& 12 \\
		Dropout rate					& 0.5 \\
		Weight decay 					& 0.01 \\
        Recurrent layer size            & 128 \\
        Input chunk size	    	    & 128 \\
        Batch       			        & 8 \\
        Gradient accumulation           & 4 \\
        Context stride                  & 16 \\
        Subsample size                  & 20, 48, 62 \\
		\bottomrule
	\end{tabular}
 }
\end{table} 

%% file: latex/Tables/corpus.tex
\begin{table}[t]
    \caption{Corpus' statistics. The paragraphs' quartiles are the documents' ones, and not on those generated after the tokenizer. All the quartile figures are rounded, and those about characters are in thousands; $Q_2$ is the median.}
	\label{table:corpus}
    \scalebox{0.73}{
	\begin{tabular}{@{\hspace{0.2em}}l@{\hspace{0.5em}}rrrrrrr@{\hspace{0.2em}}}
	    \toprule
		\multirow{3}{*}{\textbf{Class}}      & \multirow{3}{*}{\textbf{Docs.}}    & \multicolumn{6}{c}{\textbf{Quartiles}} \\
                                    &                               & \multicolumn{3}{c}{\textbf{Characters [k]} $n_c$} & \multicolumn{3}{c}{\textbf{Paragraphs} $n_p$}\\ 
                                    &                               & $Q_1$ & $Q_2$ & $Q_3$ & $Q_1$ & $Q_2$ & $Q_3$\\
		\midrule
            Amicus curiae           & 755 	& 1.5 	& 2.0 	& 9.5 	& 12 	& 13 	& 48\\
            Contract                & 998  	& 13.6 	& 37.6 	& 69.6 	& 88 	& 199 	& 367\\
            Decision                & 1,020	& 4.5 	& 16.7 	& 44.3 	& 16 	& 49 	& 134\\
            Discontinuance          & 155 	& 2.9 	& 4.6 	& 8.6 	& 22 	& 36 	& 57\\
            Expert opinion          & 587 	& 38.0 	& 70.0 	& 119.3 	& 163 	& 317 	& 685\\
            Notice of arbitr.       & 871 	& 4.1 	& 11.7 	& 34.4 	& 38 	& 83 	& 199\\
            Notice of intent        & 250 	& 7.0 	& 17.6 	& 30.6 	& 54 	& 103 	& 168\\
            Opinion                 & 937 	& 6.4 	& 15.4 	& 35.5 	& 12 	& 31 	& 73\\
            Other                   & 1,000 & 1.4 	& 1.8 	& 3.2 	& 14 	& 17 	& 32\\
            Other requests          & 248 	& 8.0 	& 27.0 	& 80.5 	& 50 	& 141 	& 436\\
            Pleadings               & 978 	& 6.1 	& 17.0 	& 45.5 	& 46 	& 96 	& 227\\
            Procedural order        & 1,000 & 2.0 	& 3.9 	& 10.8 	& 11 	& 18 	& 49\\
            Publication             & 1,001 & 3.8 	& 18.6 	& 46.3 	& 17 	& 65 	& 152\\
            Rule                    & 514 	& 17.3 	& 36.4 	& 63.1 	& 104 	& 199 	& 342\\
            Settlement agrmt.       & 143 	& 5.5 	& 11.1 	& 18.4 	& 23 	& 46 	& 105\\
            Treaty                  & 966 	& 13.8 	& 16.3 	& 21.2 	& 62 	& 73 	& 95\\
            Wiki                    & 298 	& 4.7 	& 6.5 	& 9.0 	& 17 	& 24 	& 34\\
            Witness stmt.           & 725 	& 4.0 	& 8.4 	& 21.8 	& 32 	& 54 	& 114\\
            \multicolumn{1}{r}{\textbf{Total}} & 12.4k & 3.5 	& 13.3 	& 36.7 	& 20 	& 62 & 160\\
	\bottomrule
	\end{tabular}
    }
\end{table}

%% file: latex/Tables/results.tex

\begin{table*}[t]
	\centering
	\caption{The best five models based on the dev set, and the model trained with the smallest sampling size.
	 For the results based on the test set, we present the distribution of the weighted F-score over 30 runs; $Q_2$ is the median.}
	\label{table:results}
		\begin{tabular}{crcccccc}
			\toprule
			\multicolumn{1}{c}{\textbf{Sampling}} 	& \multicolumn{1}{c}{\textbf{Additional}} 	& \textbf{Dev Weighted}	& \multicolumn{5}{c}{\textbf{Test Weighted F-score Distribution}} \\
			\multicolumn{1}{c}{\textbf{size}} 		& \multicolumn{1}{c}{\textbf{Features}} 	& \textbf{F-score}		& \textbf{Min}		& $Q_1$ 			& $Q_2$ 			& $Q_3$ 			& \textbf{Max} \\
			\midrule
			48 										& $n_p$ $n_c$ $a_{pp}$ 						& \textbf{0.896} 		& 0.883 			& 0.887 			& 0.891 			& 0.893 			& 0.896\\
			48 										& $n_c$ 									& \textbf{0.896} 		& 0.882 			& 0.888 			& 0.891 			& 0.893 			& 0.897\\
			62 										& $a_{pp}$ 									& 0.894 				& 0.886 			& 0.890 			& 0.893 			& 0.895 			& 0.886\\
			62 										& - 										& 0.892 				& \textbf{0.891} 	& \textbf{0.895} 	& \textbf{0.898} 	& \textbf{0.900} 	& \textbf{0.904}\\
			62 										& $n_p$ 									& 0.891 				& 0.875 			& 0.882 			& 0.886 			& 0.887 			& 0.890\\
			\midrule
			20 										& - 										& 0.869 				& 0.860 			& 0.867 			& 0.871 			& 0.872 				& 0.887\\
			\bottomrule
		\end{tabular}
\end{table*}

%% file: latex/Tables/detailed_results.tex
\begin{table}[t]
    \caption{Median F-score per class on the \textit{test set}. The results are concerning the best two models in the dev set (48 $n_p$ $n_c$ $a_{pp}$ and 48 $n_c$), and the best one in the test set (62 -). The median was calculated over 30 runs.}
	\label{table:corpus_results}
    \scalebox{0.84}{
	\begin{tabular}{@{\hspace{0.2em}}l@{\hspace{0.5em}}cccc}
		\multirow{2}{*}{\textbf{Class}} & \multicolumn{3}{c}{\textbf{Median F-score}} & \multirow{2}{*}
        {\textbf{Docs.}}\\	   
									& 48 $n_p$ $n_c$ $a_{pp}$ & 48 $n_c$ & 62 - & \\
		\midrule
            Amicus curiae           & 0.888	& 0.889	& 0.911	& 75\\
            Contract                & 0.934	& 0.940	& 0.934	& 100\\
            Decision                & 0.901	& 0.912	& 0.902	& 102\\
            Discontinuance          & 0.838	& 0.896	& 0.899	& 15\\
            Expert opinion          & 0.887	& 0.864	& 0.859	& 58\\
            Notice of arbitr.       & 0.811	& 0.833	& 0.845	& 87\\
            Notice of intent        & 0.602	& 0.621	& 0.636	& 23\\
            Opinion                 & 0.952	& 0.951	& 0.945	& 93\\
            Other                   & 0.911	& 0.894	& 0.905	& 100\\
            Other requests          & 0.880	& 0.851	& 0.897	& 25\\
            Pleadings               & 0.708	& 0.726	& 0.757	& 97\\
            Procedural order        & 0.931	& 0.930	& 0.945	& 100\\
            Publication             & 0.957	& 0.943	& 0.951	& 93\\
            Rule                    & 0.949	& 0.949	& 0.940	& 51\\
            Settlement agrmt.       & 0.750	& 0.750	& 0.720	& 14\\
            Treaty                  & 0.984	& 0.989	& 0.984	& 96\\
            Wiki                    & 0.982	& 0.981	& 1.000	& 28\\
            Witness stmt.           & 0.825	& 0.816	& 0.845	& 72\\
            
	\bottomrule
	\end{tabular}
    }
\end{table}

%% file: latex/Tables/inference_time.tex
\begin{table}[t]
    \centering
    \caption{Distribution of the time needed to run 30 Temporal pipelines. Each run processed 100 randomly sampled documents. Inference was done over a CPU server.}
	\label{table:inference_time}
    \scalebox{0.90}{
	\begin{tabular}{ccccc}
	    \toprule
		      \multicolumn{5}{c}{\textbf{Time [seconds per 100 documents]}} \\
            \textbf{Min} & $Q_1$ & $Q_2$ & $Q_3$ & \textbf{Max} \\
		\midrule
            467.679 & 484.853 & 498.080 & 506.678 & 533.945\\
		\bottomrule
	\end{tabular}
    }
\end{table}